\documentclass[a4paper,10pt,twocolumn,journal]{IEEEtran}
\usepackage{times}

\usepackage[
top=52.68pt,
textheight = 696pt,
textwidth=492pt,
columnsep=12pt
]{geometry}

\usepackage{multirow,array}
\usepackage{booktabs}
\usepackage{mathrsfs}
\usepackage{amsfonts}
\usepackage{array}
\usepackage{amssymb}
\usepackage{amsmath}
\usepackage{graphicx}
\usepackage{multirow}
\usepackage{amsthm}
\usepackage{color}
\usepackage[table]{xcolor}

\usepackage{epstopdf}
\usepackage{stmaryrd}
\usepackage{multirow,url}
\usepackage{cite}
\usepackage{bm}
\usepackage{indentfirst}
\usepackage{epsfig}
\usepackage{xcolor}

\usepackage{subcaption}

\begin{document}

\title{Optimization Design for Federated Learning in Heterogeneous 6G Networks}
\author{Bing Luo, Xiaomin Ouyang, Peng Sun, Pengchao Han, Ningning Ding, and Jianwei Huang\IEEEcompsocitemizethanks{
		\IEEEcompsocthanksitem
  
This work is supported by the National Natural Science Foundation of China (Project 62271434 and 62102337), Shenzhen Science and Technology Program (Project JCYJ20210324120011032), Guangdong Basic and Applied Basic Research Foundation (Project 2021B1515120008), Shenzhen Key Lab of Crowd Intelligence Empowered Low-Carbon Energy Network (No. ZDSYS20220606100601002), and the Shenzhen Institute of Artificial Intelligence and Robotics for Society. (Corresponding author: Jianwei Huang)

Bing Luo is with the Data Science Research Center and the Division of Natural and Applied Sciences, Duke Kunshan
University, Kunshan, Jiangsu, China. %This work is done when he was a Joint Postdoc at the Shenzhen Institute of Artificial Intelligence and Robotics for Society, The Chinese University of Hong Kong, Shenzhen, China, and the Department of Electrical Engineering and Institute for Network Science, Yale University, USA. 
(e-mail: bing.luo@dukekunshan.edu.cn)

Xiaomin Ouyang is with the department of Information Engineering, The Chinese University of Hong Kong, Hong Kong SAR, China (email:xmouyang@link.cuhk.edu.hk)

Peng Sun is with the College of Computer Science and Electronic Engineering, Hunan University, Changsha 410082, China. (email: psun@hnu.edu.cn)

Pengchao Han and Jianwei Huang  are with the School of Science and Engineering, The Chinese University of Hong Kong,
Shenzhen, China, and the Shenzhen Institute of Artificial Intelligence and
Robotics for Society, Shenzhen, China. (emails: hanpengchao@cuhk.edu.cn; jianweihuang@cuhk.edu.cn)

Ningning Ding is with the Department of Electrical and Computer Engineering, Northwestern University, Evanston, IL 60208 USA (email: ningning.ding@northwestern.edu).		
	}}
\date{}
\maketitle

\begin{abstract}
With the rapid advancement of 5G networks, %Recent years have witnessed a  upsurge of 
 billions of smart Internet of Things (IoT) devices along with %increasing number of devices, 
 an enormous amount of data are generated at the network edge. 
While still at an early age, it is expected that the evolving 6G network will adopt advanced artificial intelligence (AI) technologies to collect, transmit, and learn this valuable data for innovative applications and intelligent services. 
%To ease users' privacy concerns  
However,  traditional machine learning (ML) approaches require centralizing the training data in the data center or cloud, raising serious user-privacy concerns. Federated learning, as an emerging distributed AI paradigm with privacy-preserving nature,  is anticipated to be a key enabler for achieving ubiquitous
AI in 6G networks. 
However, there are several system and statistical heterogeneity challenges for effective and efficient FL implementation in 6G networks. 
In this article, we investigate the optimization approaches that can effectively address the challenging heterogeneity issues from three aspects: incentive mechanism design, 
network resource management, and personalized model
optimization. We also present some open problems and promising directions for future research.

%in AI-powered 6G networks.

\end{abstract}

\section{Introduction}
With the rapid applications of the Internet of Things (IoT), autonomous driving, industry 4.0, and metaverse, a massive volume of data is expected to generate at the network edge. The valuable data has great potential to power intelligent applications for our daily lives. %which can make incredible advances for our daily lives 
 %To analyze and exploit the large amount of data, standard machine learning (ML) techniques normally require collecting the training data in a central server, which causes huge transmission pressure (e.g., high capacity and low latency) on the current 5G networks.
While still in its infancy, it is generally believed that the Sixth Generation (6G)  systems will be established on ubiquitous artificial intelligence (AI) technologies, to enable such data-driven machine learning (ML) applications and services \cite{9237460}.  
However,  traditional ML techniques normally collect the training data in a centralized data center, which raises %causes huge transmission pressure (e.g., high capacity and low latency),   
severe privacy concerns (e.g., risk of data misuse and leakage of data owners) \cite{9146540}. 

To address the above challenge, federated learning (FL) has emerged as an attractive distributed learning paradigm (shown in Fig.~1). It enables network edge entities (clients) to collaboratively train a shared model under the coordination of a central server, while keeping the raw training data private \cite{mcmahan2017communication}. In FL, each client exploits its local dataset to compute a local model update, and the server periodically aggregates these 
local model updates to obtain a global model.  %without sharing raw data. %
%Similar to conventional distributed machine learning (DML) systems, FL lets the clients locally perform  model training and a server iteratively aggregate their computed updates. 
FL has demonstrated its success in many mobile applications (e.g., Goggle's Gboard and Apple's Siri), which makes it a high-potential enabler for AI-empowered 6G technology.

\begin{figure}[t]
	\centering
	\includegraphics[width=8.4cm]{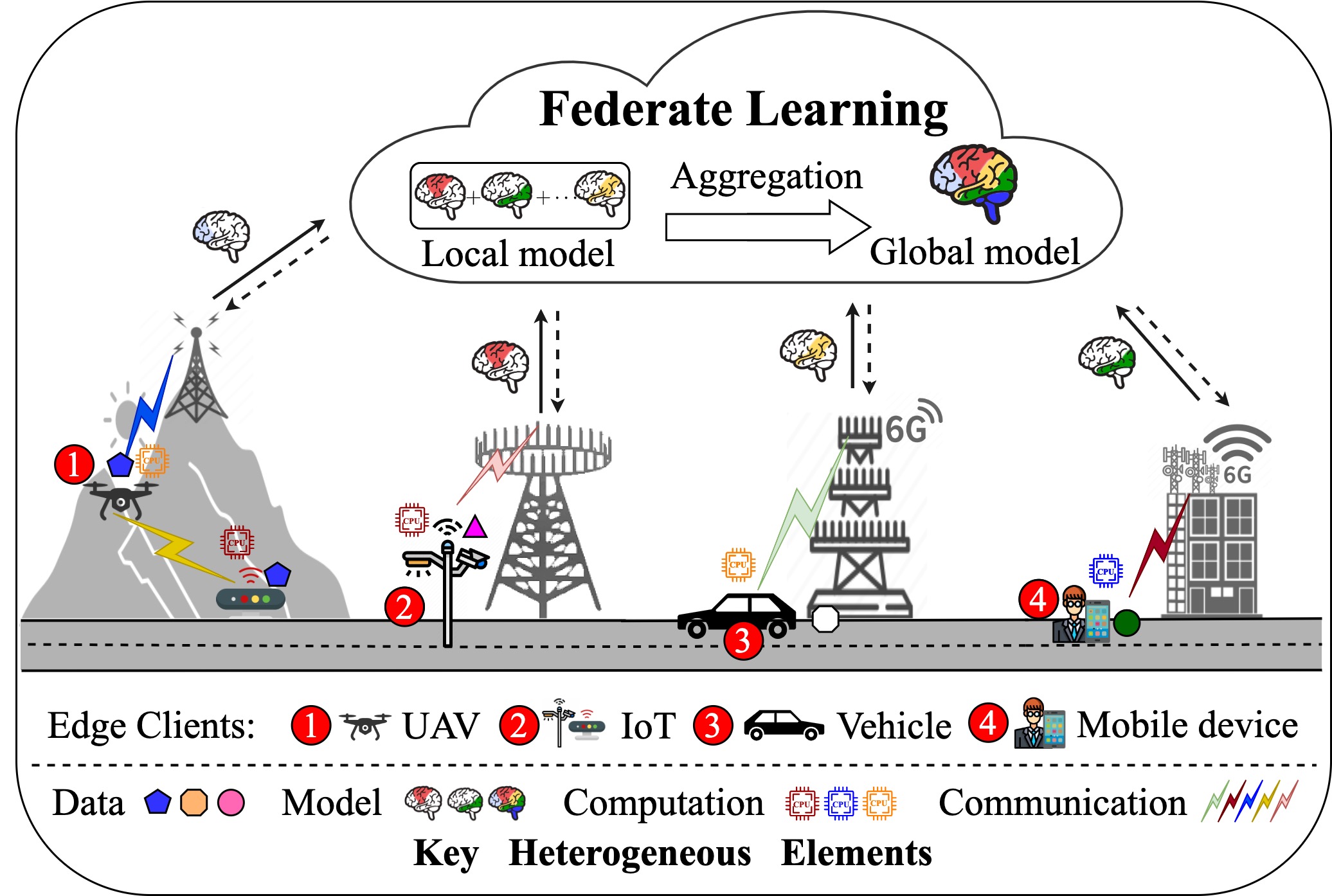}
	\caption{An illustration of FL in heterogeneous 6G networks.}\label{fig:example}
\end{figure}

%We note that the system and statistical heterogeneity are two unique features and challenges when deploy FL in current 5G networks. However, compared to current 5G networks,  6G networks is expected to  support user-centric AI with individual customized and multi-dimensional service requirement [1], where the \textbf{heterogeneity challenges will be more critical}. For example, the number of participating users will be even  large in 6G networks, with the heterogeneous resources (e.g., computation, communication and storage) will result in highly dynamic and unstable training latency for for delay-sensitive applications in 6G networks, such as autonomous driving and VR/AR games. Moreover, due to highly heterogeneous data, the demand for personalized FL model optimization will be even higher in 6G networks in order to guarantee individual’s quality of experience.

{However, the implementation of FL will face severe heterogeneity challenges \cite{li2018federated} in 6G networks. This is because unlike 5G networks that aim to improve the network performance (e.g., peak data rate and service coverage),  6G networks will be able to tailor customized services to guarantee everyone’s quality of experience (QoE). 
To achieve this goal, the 6G betwork AI architecture needs to utilize data from every user's device, and integrate heterogeneous network resources and ubiquitous intelligence from the cloud to the edge \cite{yang20226g}. Although such a cloud-edge 6G Network AI Architecture can natively incorporate FL to support user-centric AI, the individual customized and multi-dimensional service requirements will bring critical heterogeneity challenges.  
For example, massive participating users will be with highly diverse system resources (e.g., computation, communication, and storage). This can cause diverse on-device local model training latency when deploying FL, which negatively affects the application for delay-sensitive services in 6G networks, such as interactive VR/AR games.  Moreover, due to highly heterogeneous data, the demand for personalized FL model optimization will be even higher in 6G networks in order to guarantee individual’s quality of experience. }

The above characteristics and challenges will lead to effectiveness and efficiency issues for 6G FL optimization design, which can be concluded into three aspects that we proceed to discuss in this article: (i) incentive mechanism design, 
(ii) network resource management, and (iii) personalized model optimization.

%In this article, we investigate several  optimization methodologies from three typical angles that tackles both system and statistical heterogeneity for effective and efficient FL deployment in 6G networks:  %\emph{client-layer incentive Mechanism Design},  \emph{network-layer resource management}, and \emph{model-layer personalization design}. 
%\begin{itemize}
    %\item 
    
\emph{Incentive Mechanism Design}. To facilitate the optimization design of FL in heterogeneous 6G networks, we first need to design proper incentive mechanisms to stimulate sufficient client participation and contribution. Specifically, on one hand, clients participating in FL tasks incur various system costs. For example, clients sustain computation costs when computing local model updates using local CPU/GPU resources, and have communication costs when uploading locally updated model parameters or intermediate gradients. On the other hand, clients involved in FL are still susceptible to privacy threats. For example, adversaries or an honest but curious central server can infer data owners’ private information from their shared intermediate gradients or model parameters. Therefore, self-interested clients may be reluctant to participate in FL without sufficient economic compensation, which necessitates a well-designed incentive mechanism.

% \textcolor{blue}{(To Peng: please add a paragraph here, I have provide a few content from my previous survey)} [Data is the fuel of any AI application. Hence, the first and fundamental issue is obtain training data that generated in network edge devices. However, since clients and the server in FL usually belong to different entities,
% heterogeneous clients in 6G networks may have different interest and privacy constraints for the FL
% model. Hence, without sufficient compensation, clients may not be
% willing to participate due to the associated local cost (e.g.,
% resource consumption for computation and communication),
% which makes the incentive mechanism design crucial in FL
% systems.]
    
    %\item 
    \emph{Network Resource Management}. In FL, since iterative local model computation and information communications between clients and the central server can be both time and energy-consuming, it is necessary and important to analyze the incurred cost for resource-constrained edge clients. %that is incurred in a given FL task. 
    In particular, the number of participating clients is comparably large while the accessed wireless system bandwidth is limited in 6G networks. In this case,  clients may suffer from %the heterogeneous computation and communication resources will cause 
    a high transmission latency, which results in an unsatisfactory user quality of experience (QoE) for delay-sensitive applications.  %On the other hand,  reducing the number of accessing clients will result in longer rounds for reaching a target model performance  as less data is trained in each round.
    %Whereas for remote  solar-powered sensors, energy efficiency could be the major concern for the sensors to participate in FL tasks. 
    Therefore, proper network resource management %(e.g., scheduling and allocation)
    is crucial in achieving cost-effective FL in resource constrained  6G networks. %resource management (computing allocation, client scheduling, channel allocation) in these networks will be examined. Power control and management for learning and energy efficient FL
    
    %\item 
    \emph{Personalized Model Optimization}.
    % \textcolor{blue}{(To Xiaomin: please add a paragraph here to show the importance of  multi-task and personalized FL design.)}
   In a canonical FL framework (e.g., FedAvg), a central server aggregates model weights from all clients iteratively until converging to a global model. However, in real-world applications under 6G networks (e.g., smart city), the data of different clients is usually highly heterogeneous due to issues such as different user habits and physical environments. In this case, such a single model learning paradigm often suffers poor accuracy performance on the data of clients that have non-IID distributions. Moreover, there is an increasing need to improve the model accuracy on a specific user/client in federated learning. Therefore, designing a \emph{personalized FL paradigm} that can customize different models for clients with heterogeneous data during federated learning is of great significance.
%\end{itemize}

%In light of the above discussion, we %state the key contributions of this article  as follows:
%\begin{itemize}
%	\item We present three key design aspects, i.e., \emph{Incentive Mechanism Design},       \emph{Network Resource Management}, and      \emph{Personalized Model Optimization}, for optimizing FL that addresses the challenging heterogeneity issues in 6G networks.
	
%	We investigate three key FL optimization paradigms,  %from three perspectives, including 
%	i.e., \emph{Incentive Mechanism Design},       \emph{Network Resource Management}, and      \emph{Personalized Model Optimization}, in
% addressing the challenging heterogeneity issues in 6G networks.
%	\item  For each design aspect, we

In this article, we first outline the main research challenges in the above three design aspects. Then, we propose several  optimization approaches and algorithms for efficient FL deployment in 6G networks for each aspect. 
%to tackle the heterogeneity challenges in 6G networks. %optimize the effectiveness and efficiency of FL. 
Experimental results through simulated environments and hardware prototypes are also provided to validate the effectiveness of some typical research works. Finally, 
%	\item 
we identify several future research directions along with key open problems to inspire future FL research in heterogeneous 6G networks. 
%\end{itemize}

%In the rest of this article, we present the first introduce the proposed 3C framework. Then, we discuss the optimization and economic issues. Finally, we outline future research directions.

%\section{FL Optimization Framework for 6G Networks}\label{sec:3C} XXXX, In the following, we describe in details of our proposed three-layer optimization design.

%\begin{figure}
%	\centering
%	\includegraphics[width=8.4cm]{figure/2.png}
%	\caption{Proposed FL optimization framework (\textcolor{blue}{I Will ask Yifan to revise,  To all: please update the red contents)}.}\label{fig:3c}
%\end{figure}

\begin{figure}
    \centering
    \includegraphics[width =7.8cm]{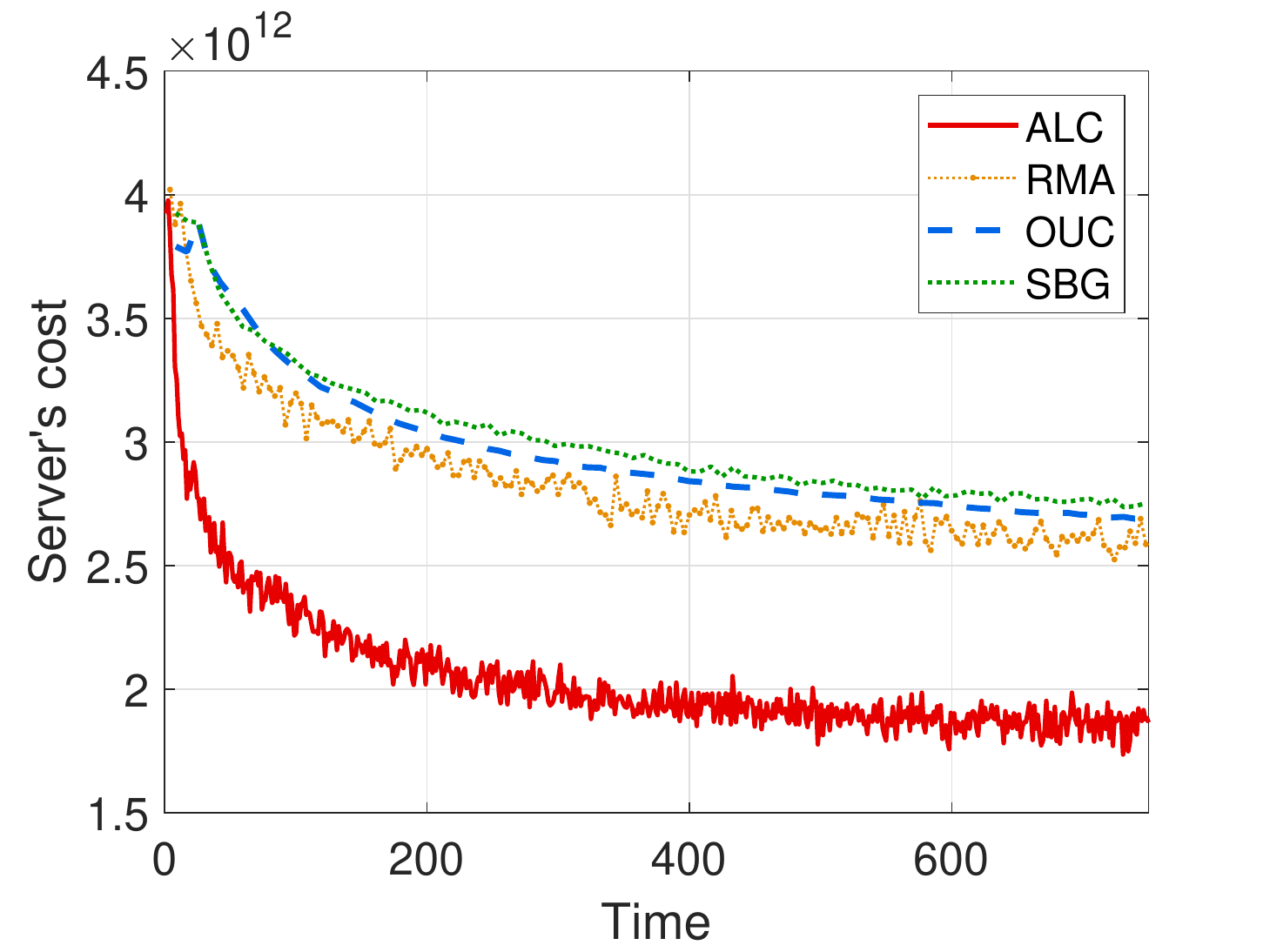}
    \caption{Server's cost of the proposed incentive mechanism on the CIFAR-10 dataset for incomplete information.}
 \label{fig:multidimen}
\end{figure}

\section{Incentive Mechanism Design for FL}
The mechanism design for incentivizing clients' participation in FL %Designing appropriate incentive mechanisms for FL 
in heterogeneous 6G networks involves several challenges. First, edge clients usually have multi-dimensional private information such as transmission delay and training costs, which are generally \emph{unknown} by the server. Thus, it is highly nontrivial for the server to selectively incentivize desirable users’ participation to enhance training efficiency and effectiveness. Second, it is challenging to simultaneously account for the privacy preservation and incentive design for clients while ensuring good FL model training performance. The reason is that there exists an intrinsic tradeoff between privacy and model performance, which is hard to analyze. Third, it is difficult to evaluate each client’s contribution in a fair and efficient manner, since the FL paradigm does not allow direct access to each client’s local data. In this section, we introduce some typical works that can well address the above challenges when designing incentive mechanisms for FL in heterogeneous 6G networks. 

\subsection{Incentive with Multi-Dimensional Private Information}
Edge clients in heterogeneous 6G networks usually have multi-dimensional private information (e.g., heterogeneous training costs and communication delay). It is necessary for the server to design an incentive mechanism to stimulate clients’ participation, encourage honest behaviors, and enhance training efficiency in the presence of clients’ multi-dimensional private information. Ding \emph{et al.}~\cite{ding2020optimal} presented an analytical study on the optimal incentive mechanism design for FL in such cases. They employed a multi-dimensional contract-theoretic approach, which summarizes clients’ multi-dimensional private information into a one-dimensional criterion that allows a complete order of clients. {They further implemented the optimal contract design under three typical information scenarios (i.e., complete information scenario, weakly incomplete information scenario, and strongly incomplete information scenario), to reveal the impact of information asymmetry levels on server’s optimal strategy and minimum cost (consisting of expected accuracy loss of the global model and the total payment to users). As shown in Fig.~\ref{fig:multidimen}, they showed that the proposed optimal incentive mechanism ALC has a much better performance compared with state-of-the-art baselines designed for non-IID data (i.e., RMA, OUC, and SBG  proposed in other literature). The maximum cost  reduction of ALC compared with benchmarks RMA, OUC, and SBG can reach 33.62\%, 36.17\%, and 37.01\%, respectively.}

% optimal incentive mechanism based on their proposed model has a much better performance, compared with the state of art in literature designed for non-IID data distributions. 

\subsection{Incentive with Privacy Preservation}
To jointly deal with clients'  privacy protection and incentive issues while ensuring satisfactory FL model training performance, researchers have developed several privacy-preserving incentives for FL~\cite{sun2022profit,sun2021pain}. For example, Sun \emph{et al.}~\cite{sun2021pain} incorporated differential privacy (DP) into FL to preserve clients' privacy. Furthermore, considering that clients under DP protection (with moderate privacy budgets) still sustain a certain degree of potential privacy disclosure and incur some privacy costs, they designed a contract-theoretic personalized privacy-preserving incentive for FL, named Pain-FL. 
% It provides customized payments for clients with heterogeneous privacy preferences as a compensation for privacy cost while ensuring satisfactory FL model training performance. 
The basic idea of Pain-FL is that the server customizes a contract item for each client, which specifies a kind of privacy-preserving level (PPL) measured by the privacy budget in DP and the corresponding payment. In each round of FL with DP, each client perturbs her calculated stochastic gradients % (to be uploaded)
with the specified PPL in her chosen contract item in exchange for the corresponding payment. They analytically derived a set of optimal contract items under both complete and incomplete information scenarios. They further empirically show that the designed incentive mechanism outperforms the uniform payment baselines in terms of the convergence error performance of the finally learned global model. 

% Besides, while bearing some desired economic properties, i.e., budget feasibility, individual rationality, and incentive compatibility. 

\subsection{Efficient and Fair Contribution Measurement}
Clients usually make heterogeneous contributions to FL model training due to factors like different training data quantity and quality (e.g., the non-IID degree of local training data). Therefore, it is crucial to accurately measure each client's contribution for fair reward allocation. Specifically, each client in FL should get corresponding rewards based on its contribution to the federation rather than the same reward, which promotes the sustainable operation of the federation. Compared to the existing contribution measurement methods that consume intensive computing resources and operate offline, Yan \emph{et al.}~\cite{yan2021fedcm} proposed a real-time contribution measurement method (FedCM) for clients in FL. FedCM defines the impact of each client, and comprehensively considers the current and previous rounds to obtain the contribution rate of each client with attention aggregation. Moreover, FedCM updates contribution aligned with FL, which enables it to implement in real-time. The authors conducted extensive experiments to evaluate FedCM, and the results show that it is more sensitive to data quantity and data quality under the premise of real-time than the state-of-the-art methods.

% \textcolor{blue}{(To Peng: please add contents here. I suggest three paragraphs:  the first paragraph can introduce the challenge in FL incentive mechanism; then use the second paragraph to describe our works (Ningning and your works), also please provide a experiment fig (yours or Ningning's paper) to highlight the results  of our incentive mechanism; After that, use another paragraph to describe one or two typical  works done by others; Note: the last two paragraph corresponds to the bullets in Fig.2. }

\section{Network Resource Management for FL} 
The resource cost for FL in 6G networks mainly occurs at edge clients' iterative local model training for computing model updates and wireless communications for transmitting model parameters, which involves both learning time and energy consumption. 
However, %due to system heterogeneity,
as the participating clients in 6G networks usually have different computational powers and wireless communications speeds, standard FL algorithm (e.g., FedAvg) may cause inefficient resource cost for achieving the required model performance, especially when the clients' data are highly non-IID and unbalanced. %In addition,   because of statistical 
In this section, we present some typical network resource management methods that can efficiently address the heterogeneity challenges in 6G networks for achieving cost-effective FL design.

\begin{figure}
	\centering
	\includegraphics[width=8cm]{	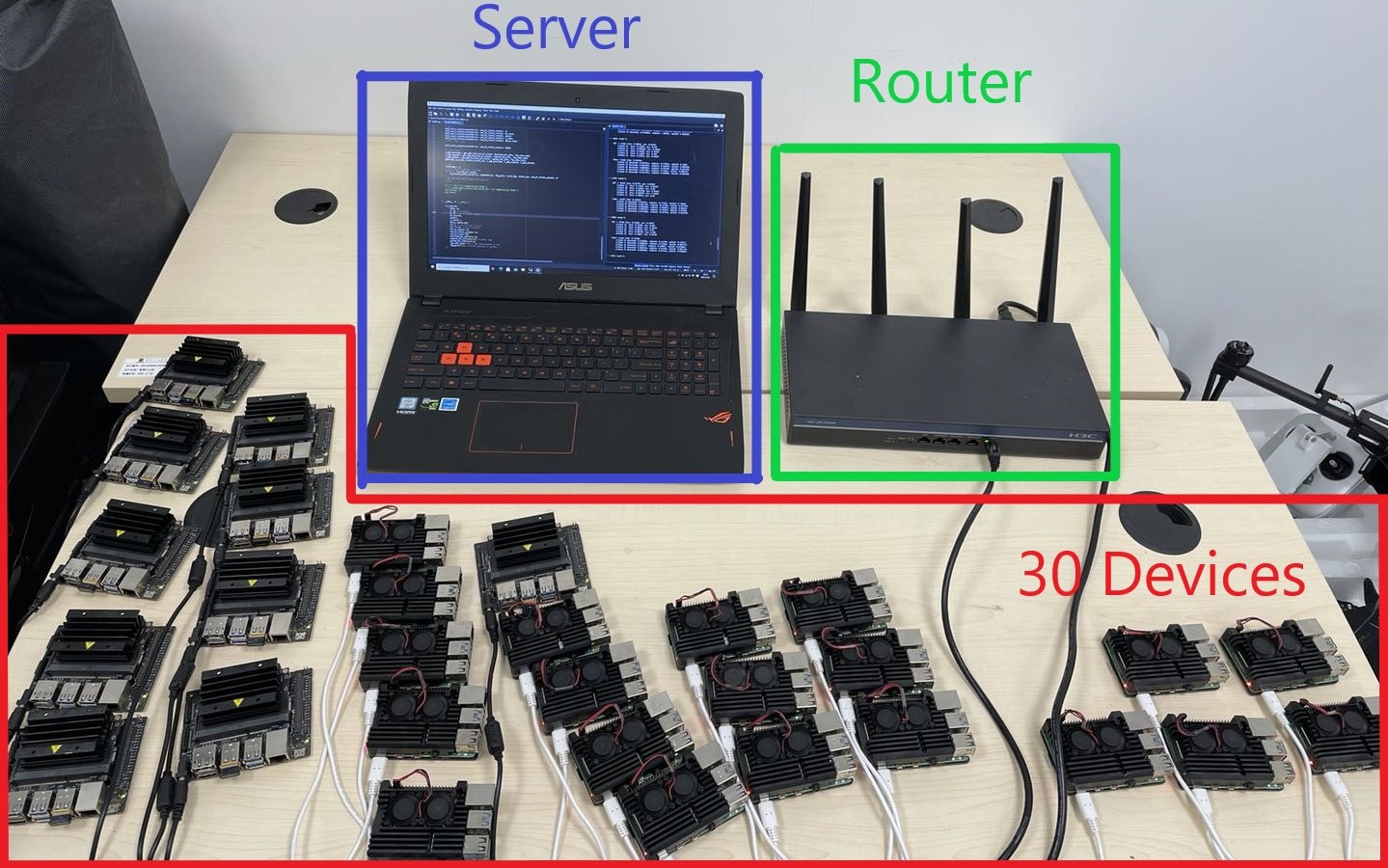}
	\caption{{Heterogeneous wireless federated learning testbed with $20$ CPU-based Raspberry Pis (version 4) and $10$ GPU-based Jetson Nanos.}} %The measured mean computation time is $4.9 \times 10^{-3}$ s with standard deviation $1.43 \times 10^{-3}$ s and the mean communication time is $0.16$~s with standard deviation $0.03$ s}}
 \label{fig_hd}
\end{figure}

\subsection{Adaptive Parameter Control}
Considering limited communication bandwidth and large communication overhead, the de facto FedAvg algorithm  \cite{mcmahan2017communication} usually performs \emph{multiple local iterations} in parallel 
on \emph{a fraction of randomly sampled clients}. These essential parameters play an important role in computation and communication resource consumption. In line with this, Luo~\textit{et al.}~\cite{9579038} studied how to design adaptive FL in wireless networks that optimally chooses these essential control variables to minimize the total cost while ensuring the required model performance. %The problem is challenging because the choices of control variables are tightly coupled due to system and statistical heterogeneity. 
The authors first analytically established the relationship between the total resource cost and the control variables with the convergence upper bound. Then, to efficiently solve the cost minimization problem, they developed a low-cost sampling-based algorithm to estimate the convergence-related unknown parameters.  %They also derived important solution properties that  helps identify the design principles for different optimization goals, e.g., reducing learning time or saving energy.
Different from most existing FL works based on computer simulations, they implement their algorithm in an actual hardware prototype with resource-constrained devices, as shown in Fig.~\ref{fig_hd}. {The developed on-device model training and real wireless communications testbed can effectively capture real heterogeneous system operation time in terms of computation and communication, which provided the design principles for FL algorithms in optimizing client sampling percentage and local iteration steps.} % which provides important design principles for FL in 6G networks.

\subsection{Importance-based Client Sampling}
\begin{figure}
	\centering
	\includegraphics[width=7.6cm]{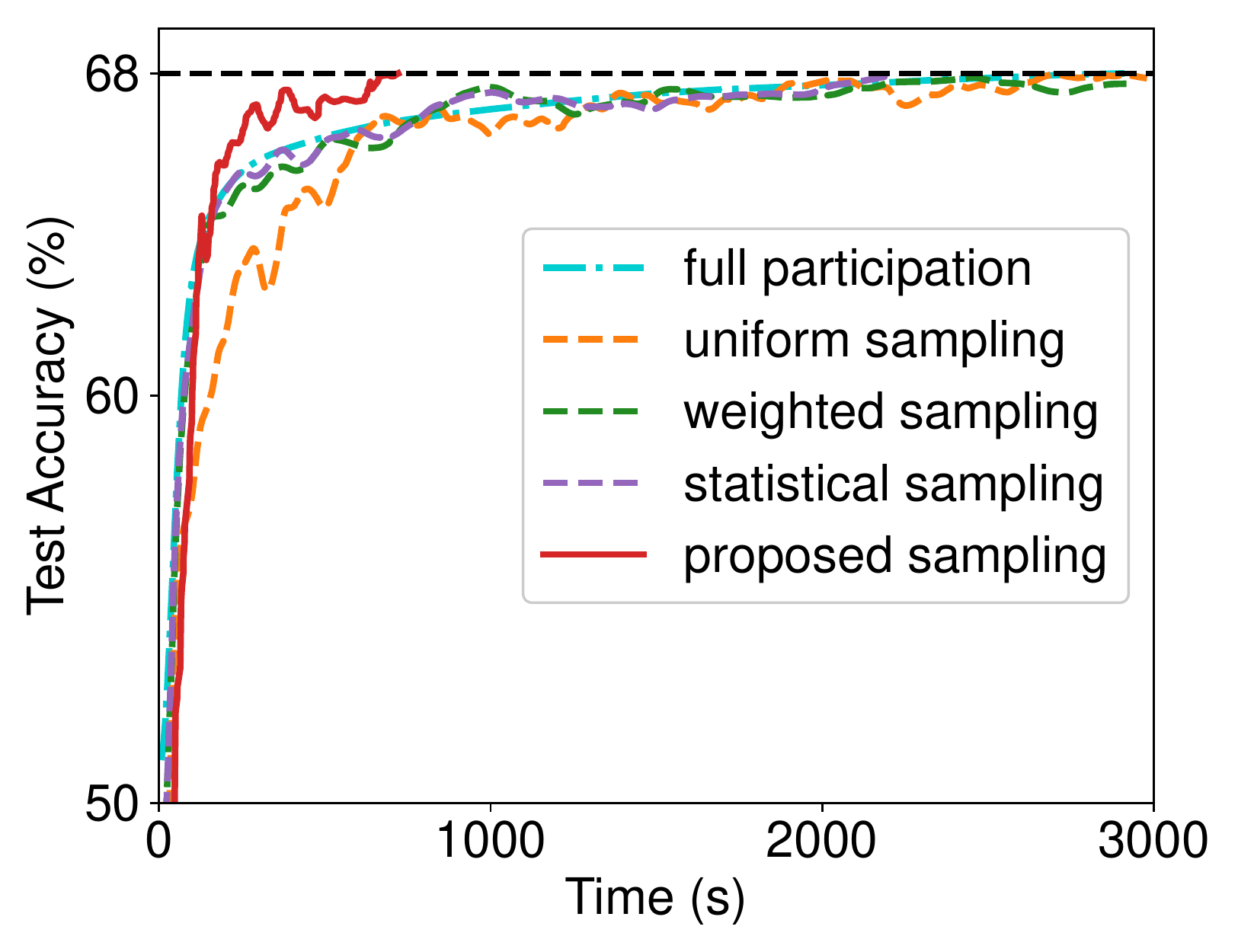}
	\caption{Testing accuracy over the wall-clock time for the EMNIST dataset on the hardware prototype using the logistic regression model.}\label{fig_sample}
\end{figure}

Existing works on the convergence analysis of FL mainly focused on 
sampling schemes that are uniformly at random or proportional
to the clients’ data sizes, which often suffer from slow error
convergence with respect to wall-clock time due
to high degrees of the system and statistical heterogeneity. To this end, the authors in \cite{9796935} proposed  an adaptive
client sampling approach 
that tackles the heterogeneity challenges to minimize the wall-clock convergence time. With an
unbiased model aggregation design,  they obtained a tractable convergence upper bound for FL algorithms with arbitrary client sampling probabilities. This allows the authors to establish the analytical relationship
between the total learning time and client sampling
probabilities and formulate a non-convex training time
minimization problem. Their solution characterizes the impact of heterogeneous computation, communication, and data distribution on the optimal client sampling probability. They also conduct experiments on a hardware prototype to validate the effectiveness of their optimized client sampling algorithm. In particular, Fig.~\ref{fig_sample} shows that for the EMNIST dataset their proposed optimal client sampling spends at least 70\% less time than other baseline sampling schemes for achieving the same target  accuracy.

%\subsection{Resource Allocation}

\subsection{Adaptive Gradient Compression}
Transmitting model parameters between clients and users in FL may lead to high communication overhead, especially for deep neural networks with millions of parameters in 6G heterogeneous networks with limited communication resources. Gradient sparsification can alleviate the communication burden of FL in 6G heterogeneous networks by only communicating a small subset of important elements of the model gradient, i.e., the top $k$ gradients with higher absolute values. In this regard,
Han \textit{et al.}~\cite{9355797} proposed an adaptive gradient sparsification approach for improving the efficiency of FL towards  heterogeneous communication resources. First, to ensure all clients contribute equally to the global model update in each round of communication, the authors design a fairness-aware bidirectional top-$k$ gradient sparsification method. Then, for given communication resource availability, the authors formulate the overall training time minimization problem to automatically determine the optimal degree of gradient sparsity (i.e., $k$). Minimizing the overall training contributes to achieving the trade-off between computation and communication. Since the system can only reveal the training time after applying a deterministic $k$, the authors propose an online learning approach to find the optimal degree of gradient sparsity using an estimated sign of the derivative of the objective function. % with a regret bound of $O\left(1/T\right)$, where $T$ denotes the number of communication round. 

\section{Personalized Model Optimization for FL} 
% \textcolor{blue}{(To Xiaomin: please discuss with Pengchao to add a few paragraphs here. I suggest three paragraphs: the first paragraph may describe the challenges; then the second may describe our works and results, also please add a figure (e.g., from your mobisys paper) to highlight the results  of multi-task FL); the third paragraph describe two typical works done by others. Note: the last two paragraph corresponds to the bullets in Fig.2. )}
% Clustering-based multi-task learning.
% Regularized local training
% Global model post-training

% Federated Learning (FL) enables training on a large amount of decentralized data residing on individual clients without sharing the raw data. 
Most FL approaches aim to learn a single model for all users, 
% by averaging the model weights on a central server
which often suffers poor accuracy performance on heterogeneous user data in real-world applications under 6G networks. The goal of personalized model optimization in FL is to customize different models for clients with heterogeneous data to improve model accuracy. There are two major challenges in personalized FL. First, there is a trade-off between the generalization and personalization ability of the learned models during federated learning, which largely affects the model accuracy. Second, federated learning on the non-IID data of clients often exhibits many convergence issues due to the divergence of model updates, which will incur significant training delays and system overhead. In this section, we introduce several representative personalized FL approaches that can customize different models for heterogeneous clients in FL.

\subsection{Clustering-based Multi-task Learning}
\begin{figure}
    \centering
    \includegraphics[width =8.4cm]{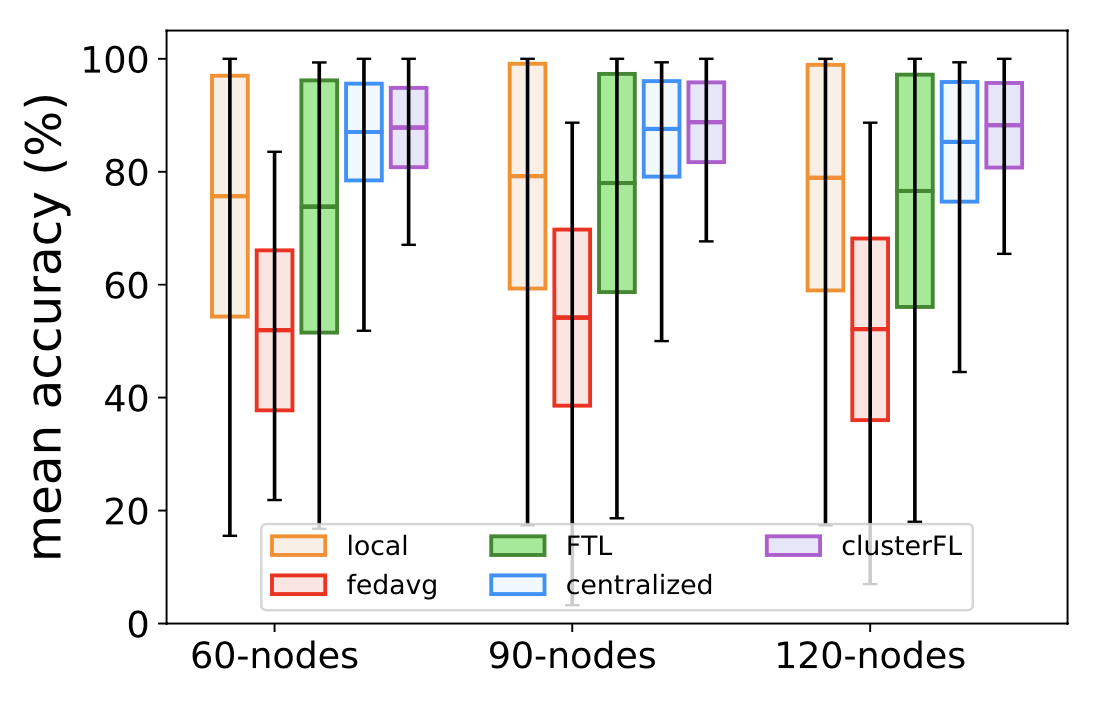}
    \caption{Accuracy performance of ClusterFL with different numbers of nodes. The results are obtained by using a large-scale dataset that collected the activity data of 121 subjects and a new FL testbed with 10 Nvidia edge devices.}
 \label{fig:clusterfl_accuracy}
\end{figure}

To train personalized models while enabling collaborative learning among similar nodes, Ouyang \textit{et al.}~\cite{ouyang2021clusterfl} proposed a clustering-based federated multi-task learning approach named ClusterFL.
% A clustering-based federated multi-task learning approach is proposed in ClusterFL \cite{ouyang2021clusterfl} to train personalized models while enabling collaborative learning among similar nodes. 
The design of ClusterFL is motivated by the key observation that the data distributions of some clients share spatial-temporal similarity in a wide range of applications with 6G networks, which can be exploited to improve the model accuracy in FL. Specifically, ClusterFL features a novel clustered multi-task federated learning formulation by introducing a cluster indicator matrix indicating the similarity of users, which minimizes the empirical loss of learned models while automatically capturing the intrinsic cluster structure among different users. 
In ClusterFL, the model weights and the cluster indicator matrix are alternatively optimized until convergence while keeping the data locality of nodes. The authors also provide theoretical analysis for achieving convergence with general non-convex and strongly convex local models.
% To solve the formulated problem, the authors propose a new distributed optimization framework based on the Alternating Direction Method of Multipliers (ADMM), which updates the model weights and the cluster structure alternatively until convergence. Moreover, to adapt the ADMM approach for the FL setting, the authors decompose the learning process into updates of local model weights to keep the data locality of nodes. 
% Besides, through the convergence analysis on the proposed federated framework, the authors provide the guidance on how to choose hyper-parameters for achieving the convergence with general non-convex and strongly convex local models.
% In ClusterFL, the authors evaluate the performance involving different numbers of nodes (i.e., 60, 90, 120) using a large-scale dataset that collected the activity data of 121 subjects. 
Fig. \ref{fig:clusterfl_accuracy} compares the accuracy performance of ClusterFL with different learning paradigms when involving different numbers of nodes. In all configurations, ClusterFL outperforms the decentralized baselines, and its accuracy even exceeds centralized learning for 60 and 90 nodes. Moreover, ClusterFL has a significantly smaller variation of accuracy among nodes, which means that ClusterFL can improve model accuracy for most nodes. 

% For the convergence performance of ClusterFL, the mean accuracy during the whole FL process increases steadily even though there are unexpected disconnections of nodes, which shows the robustness of ClusterFL.

\subsection{Regularized Local Training}
% Besides multi-task learning based FL approach that directly trains different models, other representative approaches in personalized FL includes regularized local training and global model post-training. 
Another type of personalized model optimization approach in FL is based on regularized local training. Specifically, regularization techniques are applied to limit the impact of local updates, which can provide more robust convergence performance and better-personalized models. For example, pFedMe \cite{t2020personalized} adds a regularization term to the local loss function of clients, which measures the distance between the global model and the clients' local model. Then the global model is averaged by all the local models at the central server. This approach helps decouple personalized model optimization from global model learning, thus achieving a good convergence performance. The results show that pFedMe can capture the statistical diversity of clients' data and achieves a sublinear speedup of order 2/3 for smooth non-convex objectives.

\subsection{Global Model Post-training}
There are also personalized FL approaches based on post-training of the global model \cite{yu2020salvaging}. The training process in these approaches includes two steps, federated averaging and local adaptation. Specifically, the clients will first collaboratively train a global model through federated averaging. Then the single global model learned by FL will be adapted to different clients based on their own data. The techniques for adapting the global model to clients include model fine-tuning and knowledge distillation. Therefore, the clients can individually improve the quality of their local models without re-designing the FL framework or involving other participants. However, there is a lack of theoretical analysis on how to achieve the balance of local personalization and learning from other clients for such a post-training personalized FL approach.

\section{Future Challenges and Open Issues}\label{sec:future}
In addition to the proposed  optimization methods and algorithms which address the heterogeneity issues, we further outline several future challenges and open problems in future 6G networks.

\begin{itemize}
    \item 
\emph{Incentive Mechanism for Randomized Participation.} 
Most existing incentive mechanism for FL usually assumes that all clients participate in all training rounds (known as full client participation). This assumption is generally impractical
in 6G networks due to clients' intermittent availability (e.g., unstable wireless communications or out of battery). Therefore, it is more meaningful to design a practical incentive mechanism for FL with partial client participation. However, the challenge for incentivizing partial clients is that the resulting model can be severely biased as the data on the incentivized clients 
may not be representative of all clients' data. In this case, the mechanism may fail to 
converge to the optimal model that would be obtained if all
the clients participate in training. Therefore, how to design an unbiased and convergence-guaranteed incentive mechanism requires further investigation.

\item \emph{Incentive Mechanism with Punishment Design.} Existing incentive mechanisms for FL mainly focus on how to reward clients based on their contributions. However, they have largely neglected the security issues in incentive design. Specifically, as an open and decentralized system, clients in FL can be easily compromised by external adversaries. These compromised clients can then launch Byzantine attacks via data or model poisoning to mislead the FL process and degrade the FL model performance. In such problem settings, existing incentive design will not only lead to a waste of money but also a deteriorated or even useless global model. Therefore, in order to create and maintain a benign and sustainable FL ecosystem, we may need to design incentive mechanisms that explicitly consider punishing malicious clients accordingly and induce them to behave faithfully. This is a fundamental and promising direction that is worth further investigation.

\item \emph{FL with Time-varying Data Distribution}.
In 6G networks, the clients in FL will encounter dynamic wireless channel conditions due to mobility, which leads to unstable communication rates. This problem has been widely studied in wireless communications via flexible scheduling and resource allocation algorithms. However, in addition to system dynamics,  the main dynamic challenge in FL also comes from data dynamics, where clients' local training datasets \emph{vary over time}, e.g., climate data in sensor nodes and trajectory data in autonomous cars. In the literature, 
existing FL optimization design usually assumes a fixed data distribution among
clients throughout all training rounds, which may not hold in 6G networks. The difficulty in addressing the time-varying challenge lies in the unpredictable distribution of future data, which requires new effective and robust FL learning algorithms. % to address the uncertain and time-varying data heterogeneity. 

\item \emph{Federated Knowledge Distillation}. 
{The 6G networks feature enormous number of intelligent users with customized service and QoS demands \cite{yang20226g}. The devices are normmaly with different computation capacities with limited communication resources. Dealing with these users' QoS requirements over heterogeneous infrastructures relies on a more flexible federated learning framework, including diverse model structures, different privacy protection levels, and higher model accuracy. It is possible to leverage Federated knowledge distillation (FedKD) to deal with the heterogeneous models of users and efficient communication desire, although FedKD still undergoes challenges for robustness and security design to meet the  6G version.}
% The follows red is the original and can be deleted
%\textcolor{red}{The vanilla FL algorithm (e.g., FedAvg) applies to homogeneous models with a communication overhead dependent on the model structure. Nevertheless, for heterogeneous 6G networks, the devices always have different computation capacities with limited communication resources. Federated knowledge distillation (FedKD) is a promising technique for transferring knowledge among clients using their model outputs, which always have a much smaller size than the model parameters for transmission. However, there are several challenges in achieving efficient FedKD.} 
First, FedKD is rigid in treating all clients equally while ignoring the non-IID data of clients. It is important yet challenging to find an efficient way for knowledge transfer among clients with non-IID data, other than simply taking an average of client model outputs for knowledge distillation. Furthermore, sharing the model outputs of clients to the sever faces the risk of privacy leakage. FedKD requires an additional privacy protection mechanism, which has not been stressed in the literature.

% \emph{XXXXXX}. \textcolor{blue}{(To Xiaomin: please discuss with Pengchao to add a paragraph here.)} \bigskip
\item \emph{Federated Multimodal Learning}. Most of the current studies in FL assume that there is only single-modality data on the clients. However, in many real-world applications under 6G networks such as human-computer interaction and autonomous diving, the local data on clients are usually generated from multiple modalities. Integrating information from different clients' data in federated multimodal learning has several major challenges. First, 
% there is a new dimension of heterogeneity in federated multimodal learning, i.e., modality heterogeneity. In particular, 
the local data of clients may come from multiple modalities or only a single modality due to resource limitations or sensor faults. Therefore, a scalable federated multimodal learning framework is needed to effectively aggregate the multimodal and unimodal models of clients. Second, it is more challenging to deal with the non-IID data distributions of clients in federated multimodal learning. The reason is that current personalized FL solutions are developed for unimodal settings where all clients train models with the same architecture, thus cannot be directly applied to federated multimodal learning. Therefore, it is important to design new algorithms that can improve the model accuracy of FL clients with both heterogeneous data distributions and data modalities.
\end{itemize}

\section{Conclusion}
In this article, we have investigated
three key optimization design aspects to address the heterogeneity challenges for FL in 6G networks. For each design aspect, we outline the main challenges and present several optimization approaches 
 to optimize the effectiveness and efficiency of FL. 
Both simulation and hardware prototype experiments are provided to demonstrate the effectiveness of our
proposed research works. 
Finally, we identify several future research directions along with
key challenges to inspire the future FL research.

%\section*{Acknowledgements}
%This work is supported by .

\bibliographystyle{IEEEtran}
\bibliography{ref}

\begin{IEEEbiographynophoto}{Bing Luo}
received the Ph.D. degree from The University of Melbourne, Australia. He is currently an Assistant Professor of Data and Computational Science at Duke Kunshan University, China. His research interests are federated learning and analytics, network optimization, game theory, and 5G/6G wireless communications and energy harvesting systems.
\end{IEEEbiographynophoto}

\begin{IEEEbiographynophoto}{Xiaomin Ouyang} received the B.E. degree from Xiamen University, China, in 2019. She is currently pursuing the Ph.D. degree in the Department of Information Engineering, The Chinese University of Hong Kong. Her research interests include Artificial Intelligence for Internet of Things, Smart Health, and Mobile Computing.
\end{IEEEbiographynophoto}

\begin{IEEEbiographynophoto}{Peng Sun} received his Ph.D degree in control science and engineering from Zhejiang University, China, in 2020. He is currently an Associate Professor with the College of Computer Science and Electronic Engineering, Hunan University, China. His research interests include Internet of Things, mobile crowdsensing, and federated learning.
\end{IEEEbiographynophoto}

\begin{IEEEbiographynophoto}{Pengchao Han}
 received the Ph.D. degree in communication and information systems at Northeastern University, China. She is currently a  Postdoc research associate at The Chinese University of Hong Kong, Shenzhen, China. Her research interests include wireless and optical networks, mobile edge computing, federated learning, and knowledge distillation. 
\end{IEEEbiographynophoto}

\begin{IEEEbiographynophoto}{Ningning Ding} is a Post-Doctoral Fellow with the Department of Electrical and Computer Engineering, Northwestern University, USA. Her primary research interests are in the interdisciplinary area between network economics and machine learning, with current emphasis on pricing and incentive design for federated learning, distributed coded machine learning, and IoT systems.
\end{IEEEbiographynophoto}

\begin{IEEEbiographynophoto}{Jianwei Huang} is a Presidential Chair Professor and Associate Vice President at the Chinese University of Hong Kong, Shenzhen. He has won multiple Best Paper Awards, including the 2011 IEEE Marconi Prize Paper Award. He is an IEEE Fellow and a Clarivate Web of Science Highly Cited Researcher, and currently serves as Editor-in-Chief of IEEE Transactions on Network Science and Engineering.
\end{IEEEbiographynophoto}

\end{document}